\documentclass{article}
\usepackage[preprint]{spconf}
\usepackage{amsmath,graphicx}
\usepackage{xfrac}
\usepackage{subcaption}
\usepackage[nolist,nohyperlinks]{acronym}
\usepackage{amsfonts}
\usepackage{hyperref}

\copyrightnotice{
    \copyright\ 2021 IEEE.
}

\toappear{To appear in {\it Proc.\ IEEE ICIP2021,
    September 19-22, 2021, Anchorage, USA}}

\title{Polynomial Trajectory Predictions for Improved Learning Performance}
\name{Author(s) Name(s)}
\address{Author Affiliation(s)}
\twoauthors
{Ido Freeman, Kun Zhao}
{Aptiv Services Deutschland GmbH\\
Wuppertal, Germany\\
    $<$first name$>$.$<$surname$>$@aptiv.com}
{Anton Kummert}
{Faculty of Electrical, Information and Media Engineering\\
University of Wuppertal, Germany\\
kummert@uni-wuppertal.de}

\begin{document}

    \maketitle

    \begin{abstract}
        The rising demand for Active Safety systems in automotive applications stresses the need for a reliable
        short-term to mid-term trajectory prediction.
        Anticipating the unfolding path of road users, one can act to increase the overall safety.
        In this work, we propose to train neural networks for movement understanding by predicting
        trajectories in their natural form, as a function of time.
        Predicting polynomial coefficients allows us to increase accuracy and improve generalisation.
    \end{abstract}

    \begin{keywords}
        trajectory prediction, motion understanding, autonomous driving, active safety.
    \end{keywords}

    \section{Introduction}
    \label{sec:introduction}
%    In recent years, the field of movement understanding has seen increased attention.
    Reliably predicting the future movement of road users will positively affect road safety, not only
    in completely autonomous applications but also in current features such as \ac{AEB} or \ac{ACC}.

    % For researchers, the field symbolises the progress of developments.
    Traditionally, there has been an emphasis on detection and recognition, with algorithms like MaskRCNN
    ~\cite{he2017mask} showing a good combination of accuracy and efficiency.
    However, detection is not enough for a fully functional system.
    Considering the example of an \ac{ACC} system, the detection and classification are merely the first steps in the
    pipeline.
%    With the prediction of road users' possible future trajectories, another step towards such a system is achieved.
    An \ac{ACC} system which anticipates its neighbours' movements is able to better prepare for events like
    cut-ins or sudden braking.
    % The necessity of trajectory prediction might be simple to explain in the automotive domain, yet it extends to
%    most applications in which living beings share a common space with robots.

    As with other tasks, this one was also enabled by public datasets like the well studied NGSim dataset
    ~\cite{ngsim} and the more recent, but with restrictive terms of use, HighD dataset ~\cite{highDdataset}.
%    Such datasets have shifted researchers focus onto traffic trajectories by providing an annotated playground.
    While real-world data enjoys a high credibility, it should also be used with care as it often features strong biases
    like region, weather or time of day.
%    Most agents tend to drive straight at a somewhat constant velocity and minimise interactions with other agents.
%    As an example, thanks to NGSim's sparsity,~\cite{mercat2019inertial} demonstrated near state-of-the-art results with
%    a simple Kalman constant velocity model.
    % While~\cite{coifman2017critical} provided an interesting critical analysis of the same dataset, showing several
%    problems with both labels and tracks.

%    To address sparsity, there have been works like~\cite{michelmore2019uncertainty} which used the CARLA
%    simulator~\cite{carla} to create a dataset with more interactions.
%    The benefits are clear;
%    one can generate edge cases with a larger variance, thus improving robustness and generalisation.
%    However, the artificial trajectories don't necessarily reflect tracks in the real-world, and the data tends to lack
%    naturally occurring issues like noise and wrong detections.

    Looking at the publications in the field, one sees a general agreement on the general framework for trajectory
    prediction.
    Some, like~\cite{gupta2018social}, use \acp{GAN}, others, like ~\cite{Lee_2017_CVPR}, use
    \acp{VAE}.
    There are also various attention mechanisms [\cite{Mohamed_2020_CVPR},~\cite{gupta2018social},
    ~\cite{kosaraju2019social}].
    Yet, the general structure of an encoder-attention-decoder model remains the same.
    Furthermore, the decoder predicts a series of coordinates, representing the future spatial positions of an
    agent at multiple temporal offsets, e.g., one, two and three seconds into the future.
    These offsets are constant and get hard-coded into the model during training.

    Although convenient, these predetermined temporal offsets have a few drawbacks.
    First, the predicted coordinates are predicted in parallel and are thus independent of each other.
    That is, they are not confined to any sort of temporal consistency.
    Trajectories jagged around the ground truth could potentially result in undesired side effects.
    For predicting cut-in events, for example, such jitters around a lane marker would mean multiple possible
    intersections, making it harder to anticipate the actual cut-in and thus reducing reliability.
    % Notice this could theoretically also happen with polynomials, yet it is very unlikely at the relevant
    %    polynomial degrees.
    Second, the prediction resolution of the fixed offsets is limiting.
    Certain scenarios, e.g., congested traffic, require a higher prediction resolution than a clear motorway.

    In this work, we propose a novel framework for continuous trajectory training and prediction with artificial
    neural networks.
%    It is based on the direct estimation of polynomial coefficients, defining a trajectory as a function of time.
    Our main contributions are:

    \begin{itemize}
        \item An alternative output formulation for trajectory prediction which better matches the nature of movement.
        \item An adjustment to the common uncertainty propagation to allow for a continuous variance estimation.
        \item A novel training scheme for enhanced generalisation.
    \end{itemize}
%    These main contributions are thoroughly compared to coordinate prediction and are shown to perform favourably
%    and successfully tackle the aforementioned issues.

    \section{Related Work}
    \label{sec:related}
%    The origins of the field are rooted in times when machine learning was less established.
    One of the first works in the field dates back to 1995, with Helbing et al.~proposing a model for pedestrian
    dynamics~\cite{helbing1995social} which defines attractive and retractive forces between pedestrians.
    A statistical approach for behaviour prediction in a multi-agent environment was proposed by
    ~\cite{antonini2006discrete}.
    It was later followed by ~\cite{4282823} with a Kalman filter based approach.

    More recent works mostly focus on environment understanding and the plurality of possible futures, i.e.,
    multi-modality.
    The former is represented by works as~\cite{AlahiSocialLSTM} and~\cite{deo2018convolutional} which proposed
    social pooling for multi-agent integration.
    ChauffeurNet~\cite{bansal2018chauffeurnet} modelled the environment as binary pixel masks which are
    convolved as an image.
    Finally, projects such as~\cite{bojarski2016end} train end-to-end systems for steering decisions. % This type
%    of projects has proven more interesting
%    from the research perspective than application oriented, mainly due to the lack of interpretability and reasoning
%    which are often required by regulators.

%	The second front focuses on the multimodal nature of future trajectories. Since there exist multiple plausible
%    developments of most manoeuvres, e.g. overtake from the left, from the right or slow down, predicting a single
%    modality would not provide the full information needed for reliable planing. As shown by~\cite{cui2019multimodal},
%    multimodality could be directly trained by adapting the `min-of-k` loss proposed in~\cite{rupprecht2017learning}
%    to the multimodal trajectory prediction.
%    The method of~\cite{tang2019multiple} is similar but integrates the
%    environment agents with the different modalities to formulate a distribution over agents and manoeuvres.
%	Finally,~\cite{michelmore2019uncertainty} has utilised the interpretability of Bayesian Neural Networks to define
    % a 'safety probability', a safety score for each of the possible trajectories.

    A slightly different strain of work looks into kinematic constraints.
    Instead of learning the well studied rules of physics from data, they are directly built into the
    network by design, reducing the prediction to its variable parts.
    A good overview is provided by ~\cite{rajamani2011vehicle}, while~\cite{cui2019deep} has more recently modelled
    vehicle movement using the bicycle model by predicting the acceleration and the steering angle.
%    The prediction is then the acceleration and the steering angle at a given temporal offset.

    This work proposes a loose variant of the latter class of publications.
    Instead of a kinematic model, and similar to~\cite{richter2016polynomial}, we predict continuous functions.
%    We propose a layer for the evaluation of polynomials which is supported by an error propagation layer
%    for variance estimation.
    Trained with our random anchoring scheme, we improve generalisation while being more accurate and
    flexible compared to coordinate prediction.

    Using neural networks for polynomial coefficients was discussed before.
    In 2014, Andoni et at.~predicted sparse polynomials with increased robustness using neural
    networks~\cite{andoni2014learning}.
    More recently,~\cite{perez2018learning} used polynomials for optical flow and video
    stabilisation.
%    Due to the dense nature of these domains, they seem to have encountered different challenges than polynomial
%    overfitting and the prediction of variance.
%    Both are substantial parts of our work.
    Finally, ~\cite{chrysos2020p} incorporated polynomials into their network structure.
    Their polynomial activation function in their generative model showed appealing results on a wide variety of tasks.

    \section{Polynomial Prediction Framework}
    \label{sec:framework}
    Our framework consists of three parts.
    Spatial position prediction, variance estimation and the training scheme.
    These are further discussed in this section.

    \subsection{Spatial Position Prediction}
    \label{subsec:formulation}
%    To improve generalisation and consistency of predicted trajectories, we propose to change the output's formulation.
    Assuming an arbitrary neural network, $f(\cdot)$, which takes in a time series of states, $s^i_t \in S$, of an
    arbitrary
    length, with $i \in \mathbb{N}$ different agents including the ego vehicle.
    The goal is to predict the trajectories of all agents for the next $T$ frames, with $t_0$ being
    the current time step.
%    Negative $t$ values, represent the trajectory up to the current frame.

    For normalisation, we use position increments, defining
    $\left[\delta x^i_t, \delta y^i_t\right] = \left[x^i_{t}, y^i_t\right] - \left[x^i_{t-1}, y^i_{t-1}\right]$.
    We follow the common state definition $s^i_t = \left[\delta x^i_t, \delta y^i_t, v^i_t, \alpha^i_t,
        \theta^i_t, l^i_{t}, \varphi_t^i\right]$.
    I.e., position increments, current velocity, acceleration, heading angle and, finally, the polar coordinates to
    the ego agent, respectively.
    Prior art defines the output of such a network as

    \begin{equation}
        f(S) = [x_{1}, y_{1}, x_{2}, y_{2}, \ldots, x_T, y_T].
        \label{eq:coords_func_def}
    \end{equation}
    The origin of this coordinate system is the ego vehicle at the current $t_0$.
    We define our model's output as
    \begin{equation}
        f(S) = \left[ a_1, \dots, a_{d_x}, b_1, \dots, b_{d_y} \right] = A \cup B.
        \label{eq:poly_func_def}
    \end{equation}
    Meaning that our neural network predicts two sets of parameters $A = \left[ a_1, \dots, a_{d_x} \right]$ and
    $B = \left[ b_1, \dots, b_{d_y} \right]$.
    These parameterize two polynomial functions, $x(t)$ and $y(t)$, of degrees $d_x$ and $d_y$, respectively, which
    describe the trajectory along the time dimension.
    The polynomial form for $x(t)$ is
    %	 for two functions. The predictions  parametrise the function for the $x$ coordinates while the predictions
%    are for the $y$ coordinates. % Which of these represents the lateral axis and which the longitudinal axis seems to
%    be a matter of opinion and is, practically, arbitrary.
    % We define the $x$ axis as the lateral axis while the $y$ axis assumes the longitudinal dimension. Both
%    functions take a temporal offset $t$ as a variable and output the respective predicted progress.

    \begin{equation}
        x(t) = \sum_{j=1}^{d_x} a_j * t ^ j.
        \label{eq:poly-def}
    \end{equation}

%    with $a_j \in \mathbb{R}$ being the (one of $d_x$) coefficients.
    \noindent The definition of $y(t)$ is analogous.
%    The polynomials might have different degrees.
    Furthermore, the $0^{th}$ (constant) coefficient is ignored since it represents the bias on the current position
    which is, by definition, the origin.

%    This resembles the kinematic model described in~\cite{cui2019deep}, yet our constraint is the continuity
%    of natural trajectories.
    % Similar to~\cite{cui2019deep}, we have also tried predicting the axis-wise
%    acceleration and heading angle with similar results yet a complexer system.

%    As our trajectory polynomials span the agent's coordinates over time, we can interpret the coefficients of
%    different degrees as follows
%
%    \begin{itemize}
%        \setlength{\itemindent}{2em}
%        \item[$0^\textsuperscript{th}$] means the spatial bias on the current position.
%        This coefficient is therefore ignored, i.e. set to zero.
%
%        \item[$1^\textsuperscript{st}$] Gives the linear position, in metres ($M$), as a function of time ($S$ -
%        seconds), i.e. the velocity.
%
%        \item[$2^\textsuperscript{nd}$] Following the same logic, the second degree has the unit $\sfrac{M}{S^2}$, i
%        .e. acceleration.
%
%        \item[$3^\textsuperscript{rd}$] This corresponds to the Jerk (or Jolt).
%
%        \item[$4^\textsuperscript{th}$] The forth derivative on the position is called Snap (or Jounce).
%    \end{itemize}
%
%    Terms like jerk and snap are used in robotics for planning~\cite{richter2016polynomial}.
%%    They are not always necessary, yet certain manoeuvres do seem to benefit from the higher dimensionality.
%    We find the fourth degree polynomial to best match the movement along both axes.

    \subsection{Variance Estimation}
    \label{subsec:variance_estimation}
    Joint position and variance prediction, as discussed in~\cite{AlahiSocialLSTM} and
    ~\cite{DBLP:journals/corr/Graves13}, is now a standard.
    Yet as we do not have predefined prediction points, we have to adjust the formulation.
%    This was achieved by slightly adjusting the mathematical formulation.

%    Under Gaussian assumptions, the variances for both representations, coordinates and polynomial, can be predicted
%    by the neural networks.

%    \subsubsection{Variance with Coordinates}
%    \label{subsubsec:variance-for-coordinates}
%    We assume that each position in Equation~\ref{eq:gaussian-distribution} follows a bivariant Gaussian distribution
%    and that the predictions are mutually independent.
%    The output consists of the coordinates $(x_t, y_t)$, the corresponding standard deviations
%    $\sigma_{t} = (\sigma_{x_{t}} \sigma_{y_{t}})$ and their correlation coefficients
%    $\rho_{t} = (\rho_{x_{t}} \rho_{y_{t}})$.
%    Thus giving
%
%    \begin{equation}
%        (x_{t},y_{t})\sim  N(\mu_{t},\sigma_{t}, \rho_{t}).
%        \label{eq:gaussian-distribution}
%    \end{equation}
%    The negative log probability of Equation~\ref{eq:gaussian-distribution} is then used as a loss function by
%    assigning
%    the observed trajectory to $\mu$.
%    This allows us to additionally get the gradients wrt. the standard deviation as well as the correlation
%    coefficient.

    % with $\mu = (\mu_{x}, \mu_{y})$ as the observed trajectory ground truth and negative logarithm probability of
%    Equation~\ref{eq:gaussian-distribution} as cost function for network training, the standard deviation $\sigma$ and
%    the correlation coefficient $\rho$ can also be predicted.

%    \subsubsection{Variance with Polynomials}
    Predicting polynomial coefficients renders the formulation of~\cite{DBLP:journals/corr/Graves13} inapplicable.
    We fix this by outputting the polynomial coefficients $\left[ a_{1}, \dots, a_{d_x} \right]$ along with their respective
    predicted standard deviations $\left[ \sigma_{a_1}, \dots, \sigma_{a_{d_x}} \right]$.
    Since the loss is evaluated by sampling from the position polynomials, we need to propagate the
    variances to a positional variance form.

    Following~\cite{AlahiSocialLSTM}, the respective covariance matrix
    $cov(A)$ is denoted as
    \begin{equation}
        \text{cov}(A) = \begin{pmatrix}
                            & \sigma_{a_{1}}^{2} & 0                  & \dots & 0                      \\
                            & 0                  & \sigma_{a_{2}}^{2} & \dots & 0                      \\
                            & \dots              & \dots              & \dots & 0                      \\
                            & 0                  & 0                  & 0     & \sigma_{a_{d_{x}}}^{2}
        \end{pmatrix}.
        \label{eq:coefficients-cov}
    \end{equation}

    \noindent In Equation~\ref{eq:poly-def}, $x(t)$ is a linear combination of polynomial coefficients $a_{j}$.
    Given the covariance $\text{cov}(A)$, the variance of function $x(t)$ at frame $t$ is
    \begin{equation}
        \text{var}(x(t)) = \sum_{j=1}^{d_x}\sigma_{a_j}^{2}*(t^{j})^2.
        \label{eq:variance-propagation}
    \end{equation}

    \noindent The axis-wise probability density function is then a Gaussian

%    \begin{equation}
%        P(x(t)) = \frac{\text{exp}(-\frac{1}{2}(x(t)-\mu_{x, t})^{\top}\text{var}(x(t))^{-1}(x(t)-\mu_{x, t}))
%        }{\sqrt{(2\pi)^{d_{x}}\left | \text{var}(x(t)) \right |}}.
%        \label{eq:gaussian-pdf}
%    \end{equation}

    \begin{equation}
        P(x(t)) = \frac{
            \text{exp}\left(
            -0.5 * \left(
            \frac{
                x(t)-\mu_{x, t}
            }{
                \sqrt {\text{var}(x(t))}
            }
            \right)^{2}
            \right)
        }{
            \sqrt {
                \text{var}(x(t)) * 2\pi
            }
        }.
        \label{eq:gaussian-pdf}
    \end{equation}

%    $\left | \text{var}(x(t)) \right |$ is the determinant of $\text{cov}(x(t))$.
    $\mu_{x, t}$ is the respective ground truth observation for the lateral axis at time $t$.
    Finally, the negative log probability is used as a loss and, surely enough, the term for $y(t)$ follows analogously.

%    \begin{equation}
%        loss = -log(P(x(t))).
%        \label{eq:final_loss}
%    \end{equation}

    \subsection{Training Scheme}
    \label{subsec:training}
    The polynomial output layers also allow for an improved training scheme.
%    One could try to directly optimise the coefficients by offline fitting a polynomial to each training sample.
%    Yet it would be as good as the offline curve fitting algorithm used.
%    A more robust way to train the network utilises polynomial sampling.
%    The evaluation of the predicted polynomials, following Equation~\ref{eq:poly-def}, is a mere vector-matrix
%    multiplication which has clear derivatives and is implemented in any modern deep learning framework.
    Evaluating the predictions for a series of temporal offsets requires a simple vector-matrix multiplication and
    results in a series of spatial coordinates.

    Nevertheless, training a high order, non-linear function on a few fixed offsets is the
    textbook example of over-fitting.
    As discussed in Section~\ref{sec:experiments}, when trained with two fixed anchors (coordinates at fixed temporal
    offsets),
    the model strongly over-fits the desired offsets.
    We circumvent this by introducing two adjustments to the training scheme.
    First, we increase the number of anchor points, i.e., the evaluated coordinates based on which the loss is
    calculated.
    This is a common practice with coordinates for a higher prediction resolution [e.g. ~\cite{tang2019multiple},
    ~\cite{deo2018convolutional}].
    Second, instead of fixed temporal offsets, e.g. predicting future frames $[5, 10, 15, \dots]$, we define a
    uniform distribution over an offset range $\mathcal{U}\{\text{min}\in \mathbb{N}, \text{max}\in \mathbb{N}\}$.
    For each sample during training, an offset $r$ is drawn from this distribution while all other anchor
    points are evenly spread accordingly.

    \begin{equation}
    [t_1, t_2, ..., t_T]
        = [\lfloor r * \frac{1}{T}\rfloor, \lfloor r * \frac{2}{T}\rfloor, \dots \lfloor r * \frac{T}{T}\rfloor].
        \label{eq:anchor_points}
    \end{equation}

    For example, for $4$ anchors from the integer range of $\mathcal{U}\{5, 30\}$ frames, the variate $r =
    20$ is drawn.
    The loss is calculated based on the following frame offsets $[t_1, t_2, t_3, t_4] = [5, 10, 15, 20]$.
    Non-integer frame indices are floored.
    Notice that the lower range is hence over-represented while the upper range becomes less frequent.
    To reduce this over-representation, the lower boundary of the range is set in practice to a rather large number
    while the upper boundary is set to be a bit larger than the maximal desired prediction offset.
    In the final system with a prediction range of $50$ frames, the uniform distribution is set to
    $\mathcal{U}\{35, 55\}$.

%	Being able to successfully train the polynomial prediction layer is the first of two steps. In the following
%    section, we discuss the extension of the framework to also predict the variance corresponding to the predictions.
%
%
%	\subsection{Variance Prediction}
%	\label{subsec:variance}
%	Similar to~\cite{deo2018convolutional} and most following papers, we would like to get a certainty estimation for
%    the predicted trajectories.
%    \todo{finish with Kuns help and explain variance propagation which allows us to predict the variance
%    of the coefficients instead of the coords}

    \section{Experiments}
    \label{sec:experiments}
%    We now evaluate our prediction layers using a variety of tests.
%    A visual inspection of the results is provided and accompanied by a quantitative analysis.
%    Then, the importance of our training scheme is tested.
%    Finally, we look at the performance in regard to generalisation.
%    This is analysed using both an interpolation and an extrapolation test of the predictions.
    Since our main aim throughout the experiments was to test the limits and performance of our polynomial predictions,
    all experiments were done with a simple \ac{GRU}~\cite{chung2014empirical} based encoder-attention-decoder
    architecture.
    It consists of a two-layer encoder and a three-layer decoder while the attention is based
    on~\cite{vaswani2017attention}.
    All layers have $32$ units and use the default Keras \ac{GRU} implementation.
    % Our main aim throughout the experiments was to test the limits and performance of our polynomial predictions.
    % We have, therefore, used the exact same model architecture and data structure, with the only difference being our
    % proposed prediction scheme, i.e. different output layers and a different training scheme.

    % \begin{figure}
    %     \centering
    %     \includegraphics[width=\linewidth]{figures/0.11.scene.png}
    %     \caption{
    %         Left: polynomial prediction, right: coordinates prediction. The axes are the coordinates in metres from
    %         the scene's origin. The trajectory padding represents the variance.
    %         Looking at the scene at such a scale, it is hard to notice noise and inconsistencies. These become
    %         clearer either by zooming in or by plotting the axis-wise predictions as done in Figure
    %         ~\ref{fig:axiswise_visualisation}.
    %     }
    %     \label{fig:scene_visualisation}
    % \end{figure}

    Using multi-modal architectures, we have established that modality selection, i.e., which of the modalities
    to use for evaluation, quickly become the bottleneck.
    This means that modality prediction is exchanged for modality selection for testing.
    As multi-modality is not the focus of this work, we followed~\cite{deo2018convolutional} who conditioned their
    decoder on the modality label, both in testing and training.
    % Thus obviating the issues surrounding multimodality while still enjoying the accuracy benefits of multimodal
    % systems.
    % This means, we condition our decoder on the modality label both in training and testing.
    This underlying architecture is then paired once with the polynomial framework and once with the common
    coordinates prediction framework, thus minimising external influences on the comparison.

    Due to its free licence, we used the NGSim dataset~\cite{ngsim}.
    It consists of two motorway segments in the USA, recorded by static traffic cameras at $10Hz$ and
    three different times of day (dawn, midday, dusk).
    % The data is provided as is, without a clear test, validation and train split which makes results comparability
    % somewhat challenging.
    The tracks are divided into segments of $200$ frames and temporarily split into train and test sets with a
    $3{:}1$ ratio.
    Due to the relatively small dataset, we refrained from composing a validation set.
    The training set was then filtered to half the amount of constant velocity, straight driving tracks, leaving
    us with ${\sim}7500$ training samples and ${\sim}3400$ testing samples.
    Furthermore, we always predict from the first to the last frame, i.e., no \lq{}warm-up phase\lq{} for the \ac{GRU}.
    The results in this section are either in \ac{ADE} or in \ac{RMSE} in metres i.e., lower is better.
    Time is given in seconds.

    \begin{figure}[h]
        \centering
        \includegraphics[width=0.9\linewidth]{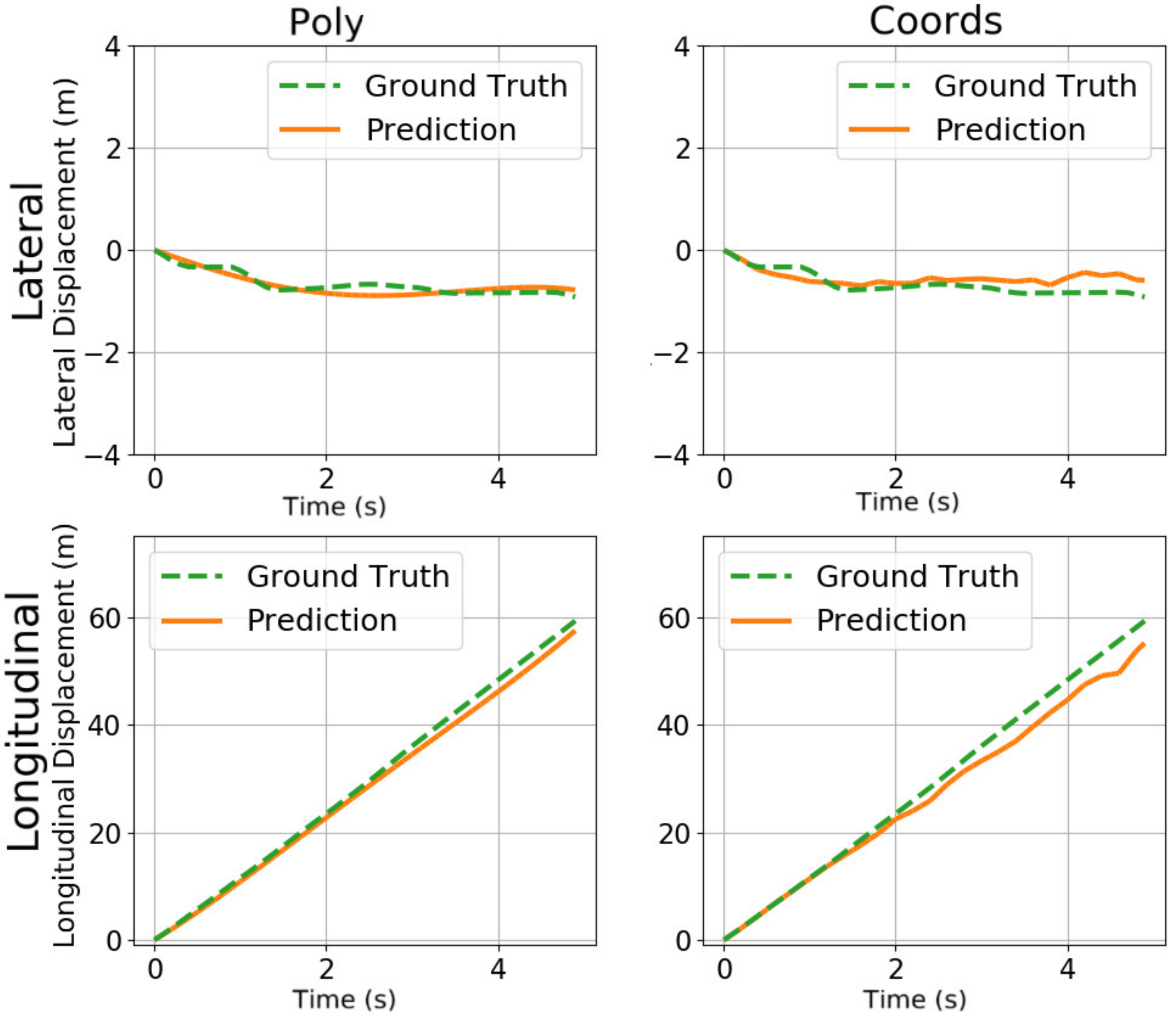}
        \caption{Axis-wise visualisation of a random prediction from the test set. Right: coordinates
        prediction. Left: polynomial prediction. The top and bottom rows show the lateral and longitudinal
        axes, respectively.
        Notice how the polynomial prediction results in smoother and less jagged trajectories.
        }
        \label{fig:axiswise_visualisation}
    \end{figure}

%    \subsection{Performance Analysis}
%    \label{subsec:performance}
%
%    This section covers the general performance analysis of our polynomial prediction model and compares it to the
%    classical coordinates prediction.
%    We open with a heuristic visual evaluation and then continue to the quantitative results.

    \subsection{Visual Evaluation}
    \label{subsec:visual-evaluation}
    Both prediction frameworks give visually appealing results.
    However, taking a closer look at a random example from the test set (Figure~\ref{fig:axiswise_visualisation}), the
    inconsistencies of the coordinates model are clearly visible.
    We explain this with two known properties of neural networks - over-fitting and prediction resolution.

    During training, the network learns to over-fit its prediction offsets, subsequently harming its generalisation
    capacity.
    This is further discussed in \autoref{subsec:random_anchoring}.

    As noticed by works like~\cite{bernardo1998regression}, neural networks can reliably regress values, but only to
    a certain precision.
    The small regression inaccuracies which create the jagged trajectories make for a small portion of the total loss,
    making them hard to correct during training.
    By training our model on continuous trajectories, we are able to reach smoother, more natural looking results, as
    seen in Figure~\ref{fig:axiswise_visualisation}.

    %	poly model used: balanced_ngsim_poly_dense_attention_17022020_0
    % 	coods model used: balanced_ngsim_coords_dense_attention_27022020_0
    \begin{table}[]
        \centering
        \begin{tabular}{p{1.1cm}p{1.15cm}p{0.7cm}p{1.8cm}p{1.8cm}}
            % \hline
            \small{Offset (sec)} & \small{Coords baseline} & \small{Poly (ours)}
            & \small{CS-LSTM (M)~\cite{deo2018convolutional}}
            & \small{MFP-1~\cite{tang2019multiple}}
            \\ \hline
            1 & \textbf{0.43} & 0.55 & 0.62
            & 0.54 \\
            2 & 1.00 & \textbf{0.93} & 1.27
            & 1.16 \\
            3 & 1.72 & \textbf{1.64} & 2.09
            & 1.90 \\
            4 & 2.76 & \textbf{2.64} & 3.10
            & 2.78 \\
            5 & 3.98 & 3.85 & 4.37
            & \textbf{3.83} \\ \hline
        \end{tabular}
        \caption{Results in RMSE of coordinates vs. polynomial training on NGSim~\cite{ngsim}. For reference, two other
        SotA results are provided.}
        \label{tab:results_comparison}
    \end{table}

    \subsection{Quantifiable Evaluation}\label{subsec:quantifiable-evaluation}
    We further evaluate our framework on the common prediction horizon of five seconds.
    The results are presented in Table~\ref{tab:results_comparison} along with two other works as a reference.
%    The results are given in \ac{RMSE} and show the improvement that our polynomial training provides over the
%    baseline.

%    only train the desired offsets, i.e., five seconds and five anchors.
%    Then, we set the number of prediction points to $25$.
%    The polynomial network is trained with the same amount of anchor points, as described in the section about
%    training.
%    Notice that this corresponds to a rather high resolution of predictions, as we actually get a prediction every
%    $0.2$
%    seconds. %Since we can't ensure that the
%    training and test sets are identical to these of other works, we provide them just as a general comparison in a
%    separate part of results table.

%    \begin{table}[]
%		\begin{tabular}{lllll}
%			\small{Offset (sec)} & \small{Coords baseline} & \small{Poly (ours)} &
%            \small{LaneGraphs~\cite{liang2020learning}} \\
%\hline
%			1       & \textbf{0.334} & 0.486 & - \\
%			2       & 0.787 & \textbf{0.767} & - \\
%			3       & 1.122 & \textbf{1.029} & - \\
%			4       & 1.274 & \textbf{1.234} & - \\
%			5       & 1.393 & \textbf{1.343} & 1.36
%		\end{tabular}
%	\caption{Results on the Argoverse dataset \todo{finalise}. The numbers on the left in each cell are the
% averages from the coordinate models while the numbers on the right are based on the polynomial models}
%	\label{fig:table_argoverse}
%	\end{table}

\begin{figure}[h!]
        \centering
        \begin{subfigure}[b]{0.48\linewidth}
            \centering
            \includegraphics[width=\linewidth]{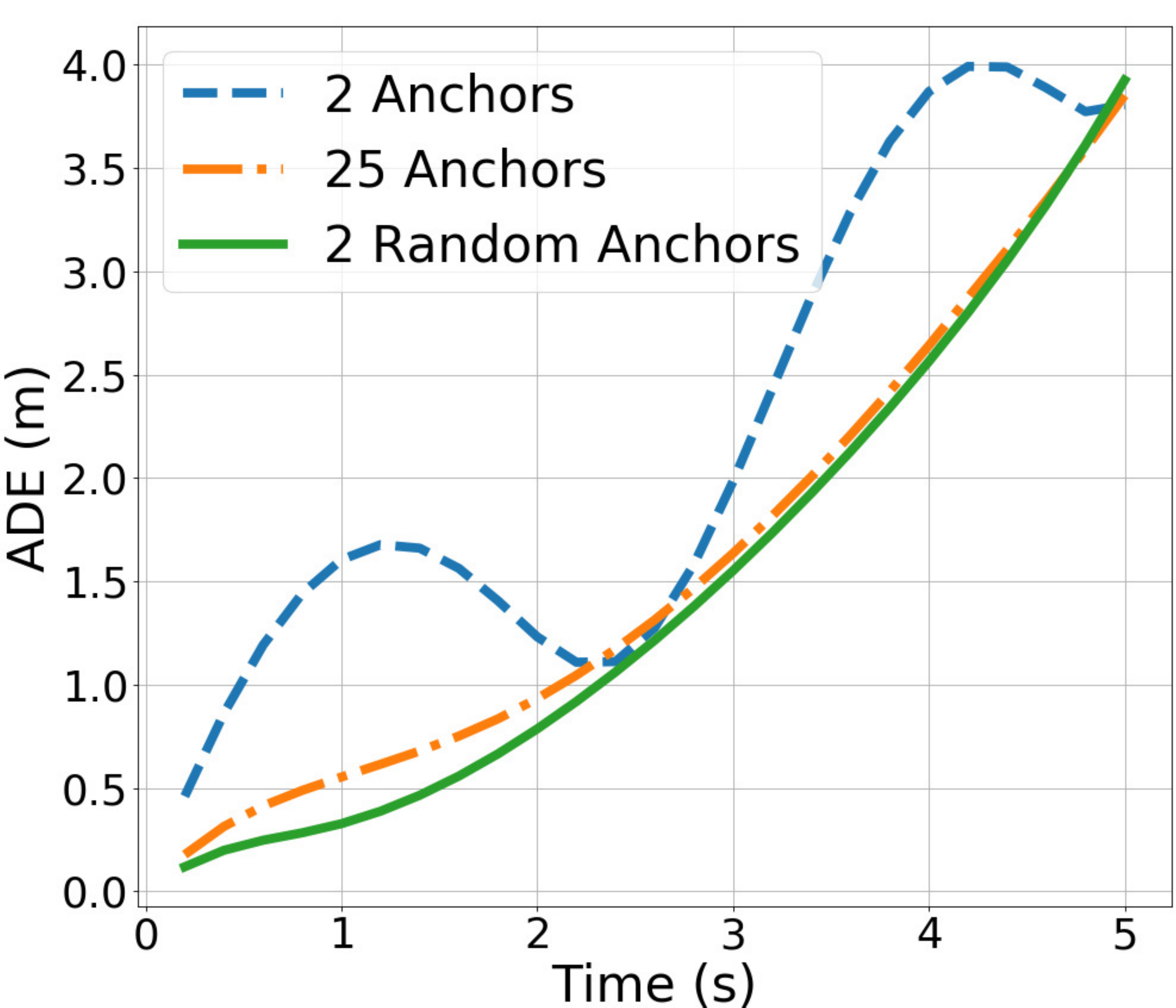}
            \label{fig:random_anchors}
        \end{subfigure}
        \begin{subfigure}[b]{0.48\linewidth}
            \centering
            \includegraphics[width=\linewidth]{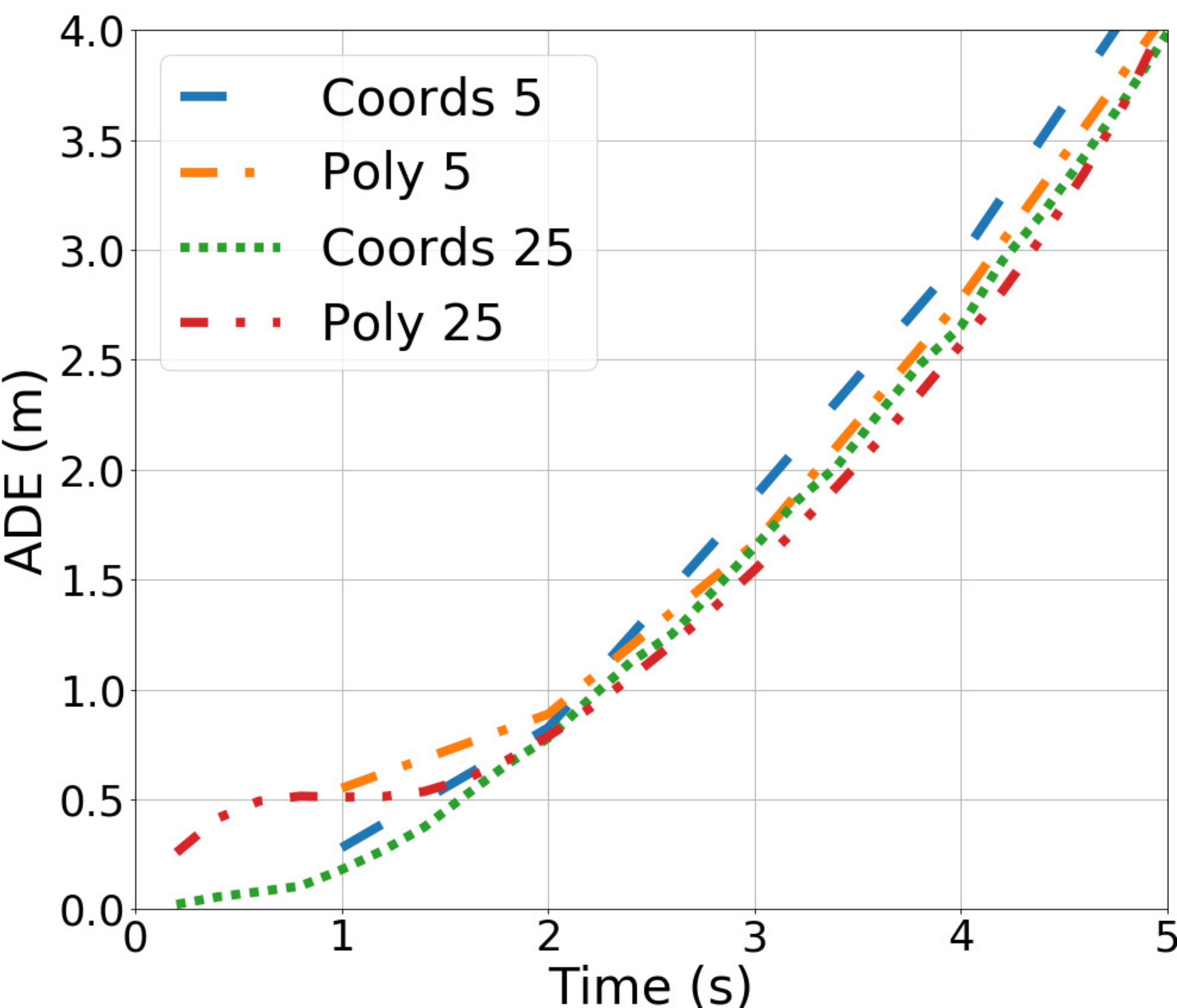}
            \label{fig:overall_results}
        \end{subfigure}
        \caption{Left: evaluating the random anchoring scheme. Even with as little as $2$ anchors the polynomial
        accuracies
        exceed the $25$ fixed anchors.
        Right: the results with $5$ and $25$ anchors per trajectory.
        For a given number of anchors, the polynomial prediction models outperform the classical coordinates.}
        \label{fig:overall_results_and_random_anchors}
    \end{figure}

    In a further experiment, we train both output types with both $5$ and $25$ anchor points.
%    The results, shown in Figure~\ref{fig:overall_results_and_random_anchors}~(right), show an edge for the polynomial
%    training.
%    Furthermore, the $5$ anchors polynomial network is almost as good as the $25$ anchors coordinates model.
    The results, shown in Figure~\ref{fig:overall_results_and_random_anchors}~(right), are quite interesting.
    Looking at the $5$ anchors and the $25$ anchors models separately, one can see favourable performance of
    the polynomial models starting at around $1.5$ seconds.
    However, even between both types of models a clear improvement is visible, supporting our claim that models better
    generalise with the increased amount of anchors.

%	\begin{figure}[h]
%		\centering
%		\includegraphics[width=\linewidth]{./random_anchors.png}
%		\caption{An evaluation of the random anchors training scheme. Notice how training with as little as $2$ random
%    anchors is partially better than $25$ fixed ones. Also notice the classical overfitting caused by training with
%    $2$ fixed anchors.}
%		\label{fig:random_anchors}
%	\end{figure}

    \subsection{Random Anchoring}
    \label{subsec:random_anchoring}

    We claim that when training on a fixed set of predetermined temporal offsets, the network only regards these few
    coordinates, thus learning less about the movement itself.
    As coordinates are fixed by definition, we test using our more flexible polynomials.
%    The results are shown in Figure~\ref{fig:overall_results_and_random_anchors}~(left).
    The first model is trained using two anchors at $t_{25}$ and $t_{50}$ frames.
    The second model is trained on $25$ evenly spread frames, i.e. $[t_2, t_4, t_6, \dots, t_{48}, t_{50}]$.
%    These two models define a range between a high over-fitting risk and a much lower one.
    The third model is trained with two random anchors as described in Section~\ref{subsec:training}.

    The results are visualised in Figure~\ref{fig:overall_results_and_random_anchors}~(left).
    Even with as little as two random labels per sample, the network manages to overcome the extreme over-fitting
    seen in the baseline and generalise better.
    \begin{figure}[h]
        \centering
        \begin{subfigure}[t]{0.48\linewidth}
            \includegraphics[width=\textwidth]{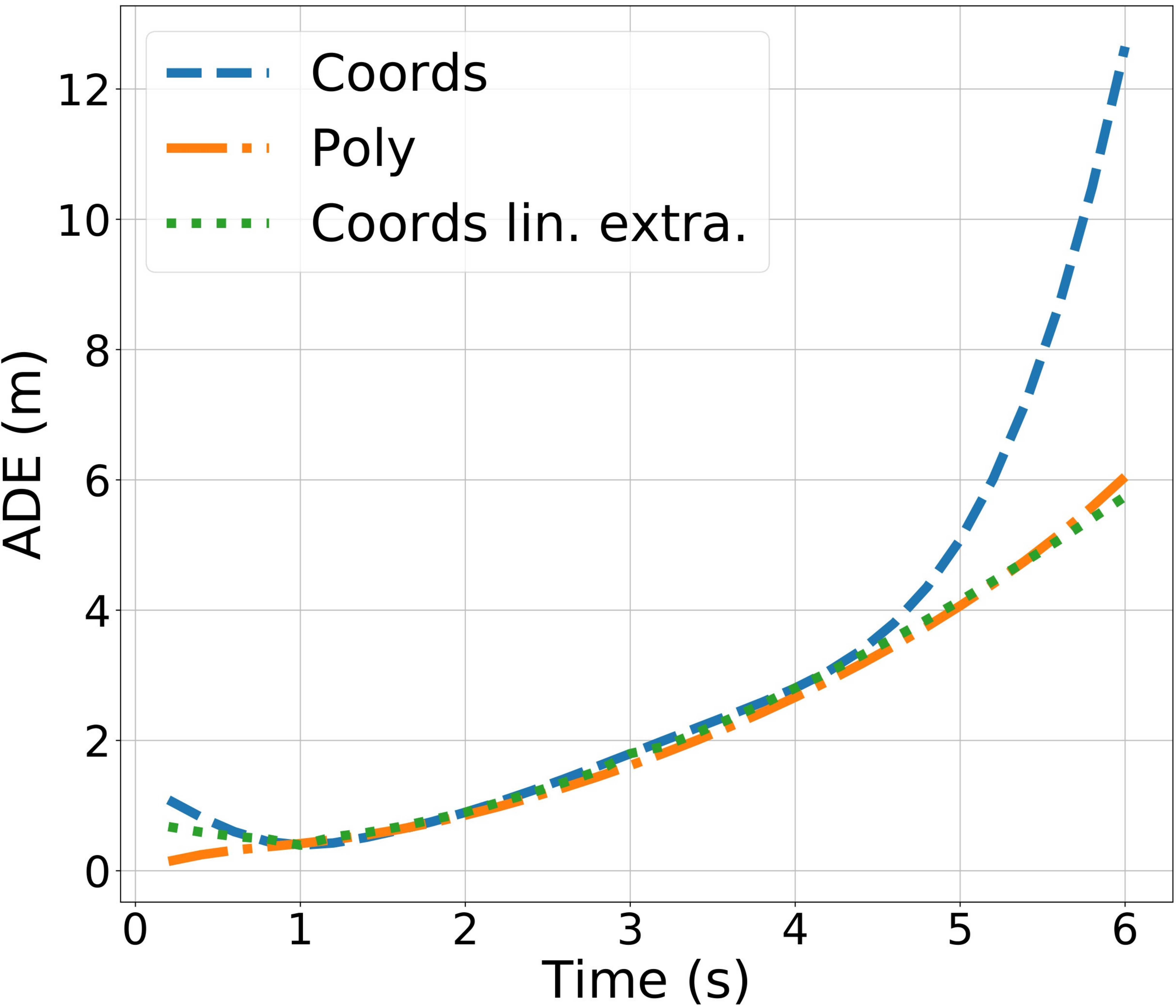}
        \end{subfigure}
        \begin{subfigure}[t]{0.48\linewidth}
            \includegraphics[width=\linewidth]{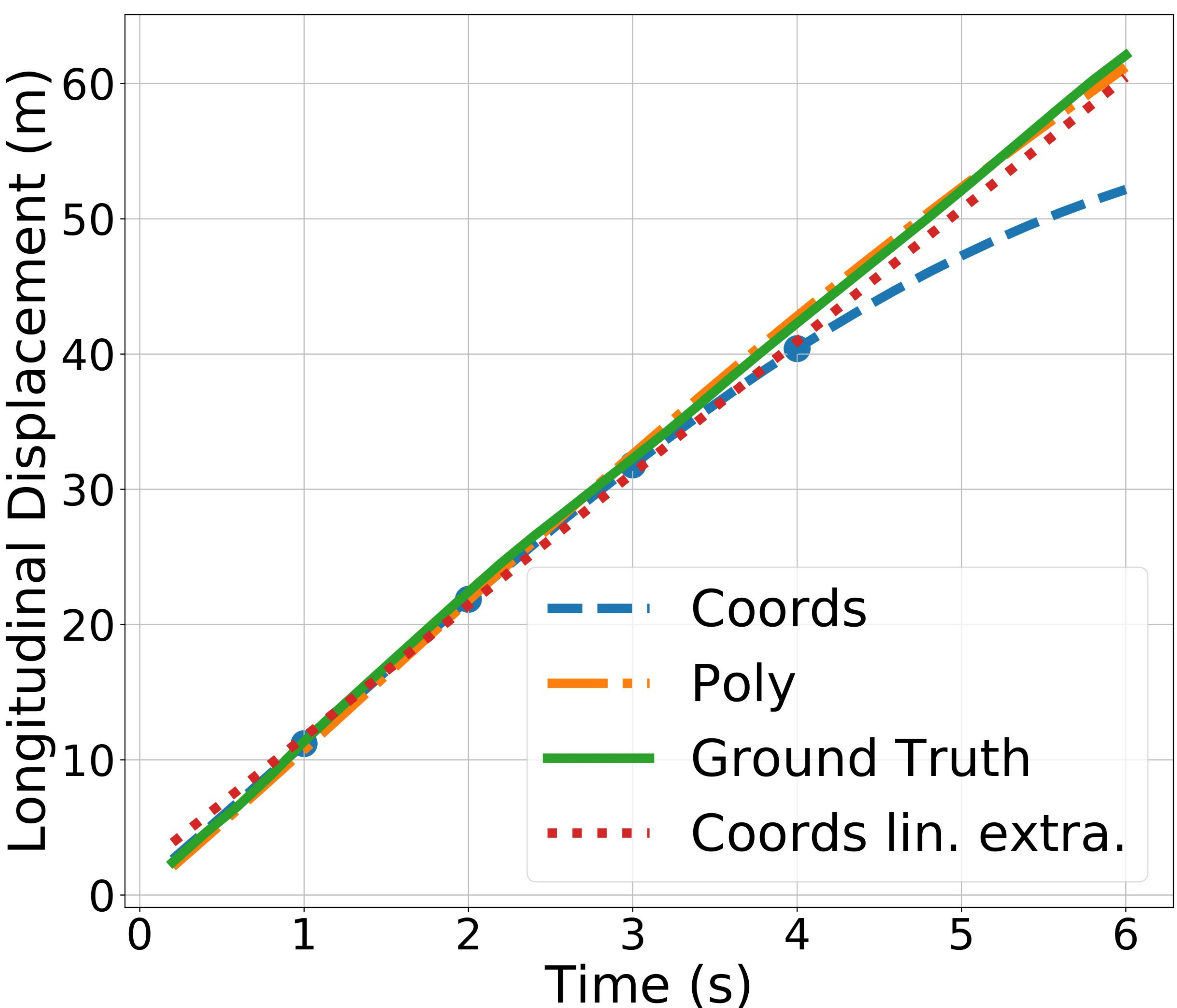}
        \end{subfigure}
        \caption{Left: the \ac{ADE} curve for the extrapolation experiment.
        Right: an example of an extrapolated longitudinal trajectory.}
        \label{fig:extrapolation}
    \end{figure}

    \subsection{Extrapolation}
    \label{subsec: extrapolation}
    To further test the generalisation capacities of our framework, we studied the extrapolation to unseen time steps.
%    Such a test does not only demonstrate the improved generalisation of our model, but also examines the effects of
%    using coordinates outside their scope.
%    For example, an application could be created to predict up to the maximal temporal offset which satisfies a certain
%    rule, e.g.~under a given variance threshold.

    A coordinates model and a polynomial one were trained on four seconds with four anchor points.
    We then evaluated them on a six seconds time span at five frames per second. % , i.e., a $0.2$ seconds resolution.
    For the coordinates extrapolation, we used NumPy's \texttt{polyfit} function and fit both a linear curve and one of
    the same order as our polynomials.

    The results averaging the entire test set are shown in Figure~\ref{fig:extrapolation}~(left).
    For the same polynomial degree our predictions better extend to new horizons.
    Yet the linear extrapolation of the coordinates prediction does similarly well.
    We use Figure~\ref{fig:extrapolation}~(right) as an example to explain this and show that often a simple linear
    interpolation performs well on average, especially on a dataset, which mostly includes straight trajectories.

    \section{Limitations}
    \label{sec:limitations}
    Despite the more accurate predictions, we have also noticed some limitations.
    For one, pedestrians typically require a higher degree of freedom than our polynomials currently enable.
    E.g., imagine a sequence of walking, stopping for a bit and then continuing to walk.
    Second, as seen in Figure~\ref{fig:axiswise_visualisation}, minor movements inside the lane are hard to represent.
    Yet such cruising artefacts are often irrelevant for most applications.
%    Finally, we were denied access to datasets like HighD~\cite{highDdataset} which would have allowed to evaluate
%    additional scenarios.

    \section{Conclusions}
    \label{sec:conclusions}
    We presented a novel lightweight framework for predicting polynomial trajectories with artificial neural networks
    by adjusting only the output layer and the training scheme.
    With the rather general constraint of continuity our framework improves the generalisation of predicted
    trajectories.
    We are furthermore positive other fields, e.g., flight path planning, could also benefit from our development.
%    Furthermore, it would be interesting to extend our polynomials to account for more complex agent movements, e.g.,
%    pedestrians.

    % We also offer a very
%    simple adjustment to the classical variance prediction layer to make it compatible with our polynomial outputs.
%    An interesting issue for future work remains the slightly worse accuracies of the polynomial prediction during the
%    first second of the trajectory.

%    \appendix

%    \section{Acronyms}
    \begin{acronym}
        \acro{ACC}[ACC]{Adaptive Cruise Control}
        \acro{VAE}[VAE]{Variational Auto-Encoder}
        \acro{AEB}[AEB]{Autonomous Emergency Braking}
        \acro{RNN}[RNN]{Recurrent Neural Network}
        \acro{CNN}[CNN]{Convolutional Neural Network}
        \acro{ADE}[ADE]{Average Displacement Error}
        \acro{GRU}[GRU]{Gated Recurrent Units}
        \acro{GAN}[GAN]{Generative Adverserial Network}
        \acro{RMSE}[RMSE]{Root Mean Squared Error}
    \end{acronym}

    \bibliographystyle{IEEEbib}
    \bibliography{refs}

\end{document}